\documentclass[letterpaper, 10 pt, conference]{ieeeconf}

\def\ps@IEEEtitlepagestyle{
  \def\@oddfoot{\mycopyrightnotice}
  \def\@evenfoot{}
}
\def\mycopyrightnotice{
  {\footnotesize IEEE-IROS-2022, Oct.23-27. Kyoto, Japan.~\copyright~IEEE All rights reserved. \hfill} 
  \gdef\mycopyrightnotice{}
}

\usepackage{eso-pic}
\newcommand\AtPageUpperMyleft[1]{\AtPageUpperLeft{
 \put(\LenToUnit{0.15\paperwidth},\LenToUnit{-1.5cm}){
     \parbox{0.9\textwidth}{\raggedleft\fontsize{10}{11}\selectfont #1}}
 }}
\newcommand{\conf}[1]{
\AddToShipoutPictureBG*{
\AtPageUpperMyleft{#1}
}
}
 
\conf{\textit{Accepted Version}: 2022 IEEE/RSJ International Conference on Intelligent Robots and Systems (IROS), Oct.23-27, 2022 - Kyoto, Japan.
~\textcopyright 2022 IEEE. Personal use of this material is permitted. Permission from IEEE must be obtained for all other uses. More information \href{http://www.ieee.org/publications_standards/publications/rights/index.html}{\textit{here}}}
\newcommand{\placetextbox}[3]{
\setbox0=\hbox{#3}
\AddToShipoutPictureFG{ \put(\LenToUnit{#1\paperwidth},\LenToUnit{#2\paperheight}){\vtop{{\null}\makebox[0pt][c]{#3}}}}
}
\placetextbox{.33}{0.05}{~\copyright~IEEE All rights reserved. IEEE-IROS-2022, Oct.23-27. Kyoto, Japan.\hfill}

\IEEEoverridecommandlockouts                              
\usepackage{epsfig} 
\usepackage{amsmath} 
\usepackage{amssymb}  
\usepackage[pdfusetitle]{hyperref}
\usepackage{subfigure}
\usepackage{multirow}
\usepackage{makecell}
\usepackage[colorinlistoftodos]{todonotes}
\usepackage{xcolor}
\hypersetup{
	colorlinks=true,
	linkcolor=black,
	citecolor=black,
}
\usepackage{soul}
\soulregister\cite7 
\soulregister\citep7 
\soulregister\citet7 
\soulregister\ref7 
\soulregister\pageref7 

\usepackage[separate-uncertainty=true,multi-part-units=single]{siunitx}
\usepackage{booktabs}
\usepackage{environ}
\usepackage{tikz}
\usetikzlibrary{matrix, chains, positioning, fit, arrows, arrows.meta, shapes, calc}
\usetikzlibrary{decorations.text, decorations.pathmorphing, decorations.pathreplacing, external}

\makeatletter
\newsavebox{\measure@tikzpicture}
\NewEnviron{scaletikzpicturetowidth}[1]{%
  \def\tikz@width{#1}%
  \begin{lrbox}{\measure@tikzpicture}%
  \BODY
  \end{lrbox}%
  \pgfmathparse{#1/\wd\measure@tikzpicture}%
  \BODY
}
\makeatother

\usepackage{array}
\newcommand{\PreserveBackslash}[1]{\let\temp=\\#1\let\\=\temp}
\newcolumntype{C}[1]{>{\PreserveBackslash\centering}m{#1}}
\newcolumntype{R}[1]{>{\PreserveBackslash\raggedleft}m{#1}}
\newcolumntype{L}[1]{>{\PreserveBackslash\raggedright}m{#1}}

\newcount\tmpnum

\def\storedataA#1{\advance\tmpnum by1
   \ifx\end#1\else
      \expandafter\def\csname data:\tmp:\the\tmpnum\endcsname{#1}%
      \expandafter\storedataA\fi
}
\def\getdata[#1]#2{\csname data:\string#2:#1\endcsname}

\title{Pedestrian-Robot Interactions on Autonomous Crowd Navigation:\\ 
Reactive Control Methods and Evaluation Metrics}

\author{Diego {Paez-Granados}$^{1\dagger}$, 
Yujie He$^2$,
David Gonon$^2$,
Dan Jia$^3$,
Bastian Leibe$^3$,\\
Kenji Suzuki$^4$,
Aude Billard$^2$
\thanks{*This work was funded in part by the EU H2020 project "Crowdbot" (779942).}
\thanks{$^{\dagger}$ is the corresponding author.
 $^1$ D.Paez-Granados is with the SCAI Lab, Swiss Federal School of Technology in Zurich - ETH Zurich and SPZ, Switzerland
    {\tt\small dfpg@ieee.org}}
\thanks{$^2$ Y. He, D. Gonon, and A. Billard are with the LASA Laboratory, Swiss Federal School of Technology in Lausanne - EPFL, Switzerland
        {\tt\small\{yujie.he; david.gonon; aude.billard\}@epfl.ch}}
\thanks{$^3$ D. Jia, and B. Leibe are with the Visual Computing Institute, RWTH, Germany
        {\tt\small \{jia; leibe\}@vision.rwth-aachen.de}} 
\thanks{$^4$ K. Suzuki is with the Artificial Intelligence Lab, University of Tsukuba, Japan        {\tt\small kenji@ieee.org}} 
 	}

\begin{document}

\maketitle
\thispagestyle{empty}
\pagestyle{empty}
\addtolength{\textfloatsep}{-0.21in}
\addtolength{\abovecaptionskip}{-0.11in}

\begin{abstract}
Autonomous navigation in highly populated areas remains a challenging task for robots because of the difficulty in guaranteeing safe interactions with pedestrians in unstructured situations.
In this work, we present a crowd navigation control framework that delivers continuous obstacle avoidance and post-contact control evaluated on an autonomous personal mobility vehicle. We propose evaluation metrics for accounting efficiency, controller response and crowd interactions in natural crowds. 
We report the results of over 110 trials in different crowd types: sparse, flows, and mixed traffic, with low- ($<0.15\ ppsm$), mid- ($<0.65\ ppsm$), and high- ($<1\ ppsm$) pedestrian densities. 
We present comparative results between two low-level obstacle avoidance methods and a baseline of shared control.
Results show a 10\% drop in relative time to goal on the highest density tests, and no other efficiency metric decrease. Moreover, autonomous navigation showed to be comparable to shared-control navigation with a lower relative jerk and significantly higher fluency in commands indicating high compatibility with the crowd.
We conclude that the reactive controller fulfils a necessary task of fast and continuous adaptation to crowd navigation, and it should be coupled with high-level planners for environmental and situational awareness.
\end{abstract}



%
\IEEEpeerreviewmaketitle

\section{Introduction}
\begin{figure}[!t]
    \centering
	\includegraphics[width=8.4cm]{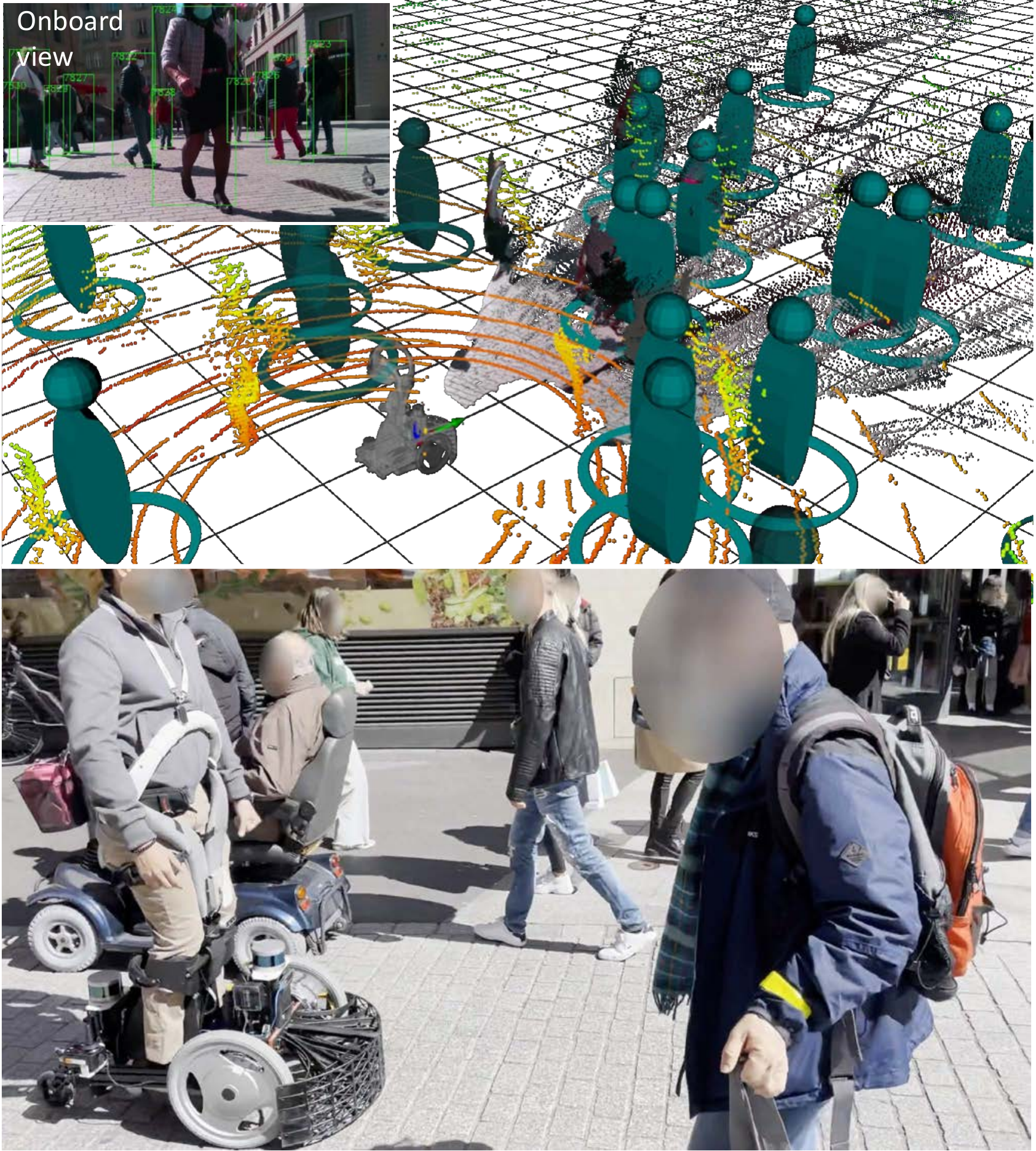}
	\caption{Crowd navigation evaluation around the city of Lausanne, with the robot Qolo using reactive navigation for obstacle avoidance and post-contact compliance. Top, depicts the on-board people detection and tracking from Lidar and RGBD data. Bottom, navigation view from the external camera in dense mixed crowds environment}\label{fig:intro}
\end{figure}

Mobile service robots offer great societal value, such as transporting personal mobility devices (Segway, USA, Rokuro, Japan), last-mile delivery services (Starship Inc. USA), autonomous cleaning robots (Bluebotics, Switzerland), autonomous wheelchairs (Whill Inc. Japan), telepresence robots and tour-guide robots.
Nonetheless, most robots are still limited to navigation in low-density areas, such as pathways or large open areas with the basic safety control system setting the robot to freeze as soon as it perceives a likely contact with pedestrians \cite{Trautman_Patel_2020, Sath2020}. Such a reaction would most likely be unexpected by pedestrians and lead to more dangerous collisions with pedestrians stumbling on the robot, a "frozen" robot would become a danger to itself and bystanders \cite{Salvini2021-THRI,Salvini2021-soro}. This is the case, in highly dynamic environments such as malls, airports, markets, and mix-traffic zones with other mobility vehicles (as shown in Fig. \ref{fig:intro}). 

However, guaranteeing obstacle avoidance during navigation in highly occupied areas would be unattainable for current mobile service robots bounded by actuation power, computational resources, and expected to behave as pedestrians, i.e., holonomic, reactive, communicative, and knowledgeable of proxemics and other social rules.
In this work, we investigate possible reactive control strategies that avoid the "freezing" robot problem and experimentally validate the compatibility and interaction with bystander pedestrians.

Different control approaches have been proposed for dynamic obstacle avoidance, from model predictive control (MPC) applied as a quadratic program for avoiding collisions (combining braking and steering) \cite{Yi2016}, or computing interactions with other agents and formulating crowd navigation as an optimal control problem \cite{Chen2021}. As well, the partially observable Markov decision process (POMDP) is a common framework for planning with other agents' uncertain intentions ~\cite{Bai2015, Luo2018, Cai2021}. 
Other works in path planning solve the dynamic environments through models of the crowd trained in simulation. The work in \cite{Sath2020} uses deep reinforcement learning to plan the robot's trajectory through a surrounding crowd's motion which is tracked and predicted by a separate algorithm beforehand. \cite{Sath_ral_2020} works similarly with predictions of zones where the robot could become obstructing or freeze, and the approach then avoids these zones. While the work in \cite{Kobayashi2021} presents a time efficiency-based method for path planning in crowd environments.
Nonetheless, none of these works guarantees continuous operation of the robot navigation or offers a solution for fully blocked passages, likely to occur in dense crowds. Moreover, none of these methods has been systematically evaluated in natural crowds.

Dynamical systems (DS) based obstacle avoidance provides a fast and continuous solution through modulation of obstacles, as offered in \cite{Huber2019}. We have augmented this approach with a compliant control mechanism, herewith mitigating risks and allowing the robot to navigate in post-contact scenarios through sliding behaviour, as proposed in \cite{Paez_icra22}. This approach offers the reactivity of time-invariant DS combined with contact estimations of impacts through a compliant bumper, herewith ensuring impact absorption through passive and active compliance for mitigating unexpected impacts with mobile robots.

In this work, we provide a framework for crowd navigation through an integrated reactive controller for obstacle avoidance and post-contact sliding control.
We contribute by providing a systematic evaluation methodology and metrics for robot assessment on crowd navigation tasks. We present the results from testings in multiple natural crowd densities: low ($<0.15\ ppsm$), medium ($<0.65\ ppsm$), and high ($<1\ ppsm$), and crowd types: sparse, flows and mixed traffics.

Previous works on crowd navigation have focused on three main metrics: collisions, success rate, and time to goal \cite{ChenC2019, Kobayashi2021}. While the work in \cite{Kobayashi2021}, proposed a mid-density flow evaluation ($<0.5 ppsm$) with volunteer participants in a controlled setting, and used time efficiency and boundary violations (virtual collisions) as main metrics.
In contrast, we propose evaluations on natural crowds with additional metrics that assess path efficiency, controller performance, and robot-crowd interaction, similar to the simulation framework offered in \cite{Grzeskowiak2021}.

We validated the whole architecture for crowd navigation and three reactive navigation controllers on a personal mobility vehicle - Qolo \cite{Paez_tmech_2022} shown in Fig. \ref{fig:intro}, a type of powered wheelchair for standing mobility of lower-limb impaired people.
Although results in this work were shown on a person carrier robot, the proposed controller and metrics are equally valid for any mobile robot.
We provide the whole dataset of the current experiments as open-access \footnote{Dataset website: \url{https://www.epfl.ch/labs/lasa/crowdbot-dataset/}} in \cite{Paez_CrowdDataset_2021}. As well as, all source code for processing and analyzing interactions
\footnote{Data analysis tools can be found here: \url{https://github.com/epfl-lasa/crowdbot-evaluation-tools}}.
Which should enable future research on understanding people's navigation around robots, and improving detection and tracking methods.


\section{Problem Statement} \label{sec:problem}
Obstacle avoidance methods usually consider a bi-state problem with collisions as an absolute negative state which in turn leads to the “freezing robot” problem \cite{Trautman_Patel_2020,Sath2020}. 
Nonetheless, contact might be inescapable when the robot's kinematic and dynamic constraints are below the pedestrians. Hence, collisions might become unavoidable even in simple scenarios. 
In our previous work \cite{Paez_icra22}, we focused on combining active compliance with DS-based obstacle avoidance which provides a way to slide around obstacles while in contact and continue moving towards the goal. Sliding in contact limits force to a determined safe threshold following safe design considerations for robot impacts \cite{Paez-NSR-2022}.

In this work, we target to prove the feasibility of crowd navigation in natural crowds through reactive navigation control and developed appropriate metrics for its assessment.
We compare two methods for obstacle avoidance and compared their performance with shared control (SC) as a baseline with user-given commands \cite{Chen2020}.
We explored three different controllers focusing on the following questions:
\begin{enumerate}
    \item What difference can be observed in the representation and control of an autonomous robot around actual crowds?
    \item How do the proposed reactive navigation methods perform when the crowd density changes?
\end{enumerate}

\subsection{Control Architecture for Reactive Control in Dense Crowd Navigation}\label{sec:controller}

\begin{figure}[!t]
    \centering
    \includegraphics[width=8.5cm]{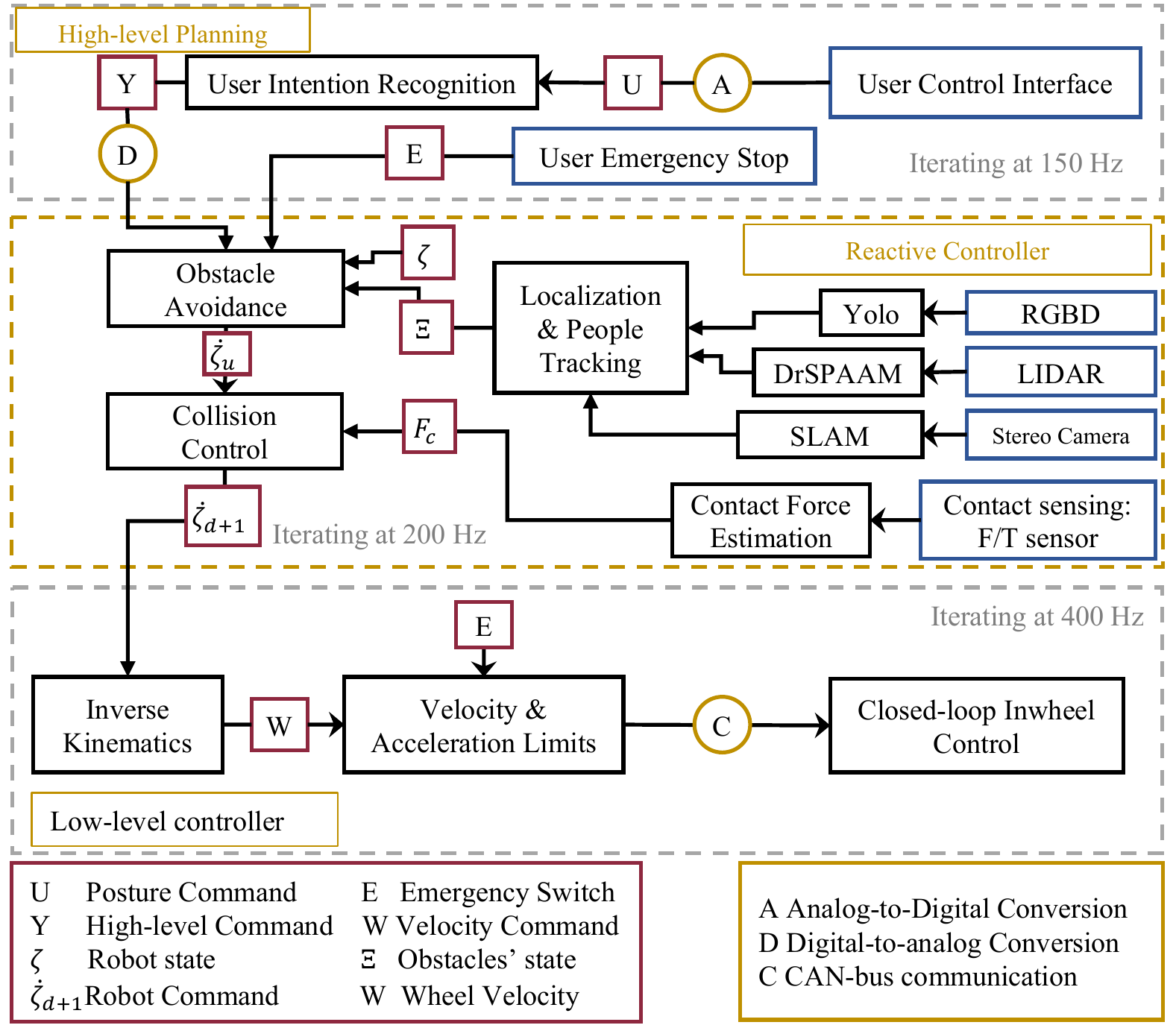}
    \caption{Controller architecture proposed for reactive navigation control on crowds navigation. High-level planning was set from the user in shared-control low-level blending or full autonomy it was replaced by a linear DS towards an attractor at the desired location (20 to 50m ahead). Obstacle avoidance used a VO-based optimization \cite{Gonon2021} or a modulated Dynamical System (DS) \cite{Huber2019}. Finally, a passive DS handles the post-collision through sliding \cite{Paez_icra22}. \label{fig:controller}}
\end{figure}

In this work, we formulated the crowd navigation control by dividing it into three main layers (as depicted in Fig. \ref{fig:controller}): 
First, the high-level planner decides the motion direction and velocity. In the robot Qolo a user intention recognition for shared control is possible through a hands-free user interface \cite{Chen2020}, or a standard virtual joystick.
Second, a reactive control layer that deals with local obstacle avoidance and post-collision control through closed-loop dynamics of the robot in its control space ($\dot{\xi}_u$). 
The main objective of the reactive control module is to provide a layer for local navigation assistance and immediate responsiveness to unmodeled or unpredicted object occurrences or motions. Therefore, this algorithm assumes a continuous dynamical system (DS) guiding the robot's motion to be existing and given by a high-level algorithm that optimally plans towards its intended goal (a user-given command, or a DS from planners).

For obstacle avoidance, we use a Modulated Dynamical System (MDS) \cite{Huber2019}, representing obstacles analytically as star-shaped level sets of a distance function that absorb the robot's footprint, thus, allowing the robot to be represented as a point moving in Cartesian space. In the case of Qolo, we control a single point in the bumper area as a holonomic point. It guarantees to lead the robot to its goal (hereafter called attractor) by assuming a circular virtual boundary (as depicted in Fig. \ref{fig:robot_setup}). Further details of this formulation are described in \cite{Paez_icra22}.
Similarly, local obstacle avoidance could be achieved through the Reactive Driving Support (RDS) proposed in \cite{Gonon2021}, which replicates the behaviour of the MDS locally through Velocity Obstacles (VO) based controller. While it sacrifices the guarantee to reach the global goal (by itself, i.e., without an additional path planner), it allows representing more accurately the robot shape detail and the non-holonomic constraints of the robot's kinematics.

For collision control, we have proposed a compliance and contact force controller through a sliding method using a known sensing surface over the robot's hull with a limited contact reference force $F_n$. Ensuring that the robot reacts to unexpected contacts (in this case, limited to the frontal bumper) and advances with a sliding manoeuvre should the underlying obstacle avoidance lead away from the contact surface without colliding with other obstacles, as we proposed in \cite{Paez_icra22}.

We modelled the robot dynamics as:$M\ddot{\xi} + C\dot{\xi} = \tau_c + \tau_e$, where $\dot{\xi} \in \mathbb{R}^2$ represents the robot's Cartesian velocity as a time invariant position dependent dynamical system. $M \in \mathbb{R}^{2\times2}$ corresponds to the virtual mass of the robot, $C \in \mathbb{R}^{2\times2}$ accounts for centrifugal and Coriolis terms, $\tau_c$ represents the control forces and $\tau_e$ any external disturbances to be rejected.
We use a controller of the form:
\begin{IEEEeqnarray}{rCl}
\tau_c &=& \lambda_t f_u(\dot{\xi}) + (F_n + F_c) \hat{n} - D \dot{\xi}
\ ,
\end{IEEEeqnarray}
where $f_u(\xi)$ represents the driving force generated by the nominal DS from the obstacle avoidance input, applied tangentially to the collision surface ($\lambda_t$) in case of contact.
$F_n $ represents a chosen force limit bounded by safety and acceptability, while $F_c $ represents the estimated contact force in the $\hat{n}$ normal surface. 
$D \in \mathbb{R}^{2\times2}$ represents a negative defined damping effect used for controlled sliding during contacts.
We transform the velocity to the domain of the robot by a first-order Taylor expansion, thus the control for the desired velocity $\dot{\xi}_{d}$ is, 
\begin{IEEEeqnarray}{rCl}
\dot{\xi}_{d+1} &=& \frac{T_s}{M} \left(
    (F_n + F_c) \hat{n} - D \dot{\xi}_d
\right) + \hat{t}^T \dot{\xi}_u \hat{t}
\ ,
\label{eq:vel_cnt}
\end{IEEEeqnarray}
where a discretizing time constant ($T_s$) is set to the sampling time of the control loop, and $M$ is a virtual desired mass for impact response.

The third and final layer is a low-level control that guarantees closed-loop velocity control and state feedback to the higher-level controllers.

A key component for crowd navigation is the localization and tracking of the environment and crowd ($\Xi$). We have set people tracking through a real-time pipeline of sensing fusion of Lidar-based detection by DR-SPAAM \cite{Jia2020} and RGBD detection by YOLO \cite{yolo_2021}. Whereas robot state ($\xi$) was estimated on optical flow principle from a stereo camera (Intel T265) fused with IMU and odometry.
The full controller repository can be found in \cite{Paez_Qolo_github}.

\section{Evaluation Method for Crowd Navigation} \label{sec:methods}

\subsection{Evaluation Metrics}
Evaluation of performance in crowd navigation for mobile robots should account for the robot's performance in achieving its task, as well as the compatibility with pedestrian navigation response and safety-related metrics.

We integrate and present feasible metrics for crowd-robot interaction from embedded sensing and perception pipelines. Moreover, we make use of 3D point-cloud sensing for further assessment of the robot's navigation performance in the crowd through 3D people detection \cite{Jia2021}, and tracking \cite{Weng2020_AB3DMOT}.
The following metrics were considered:

\subsubsection{Path Efficiency}

\textbf{Relative time to goal}: The relative time to goal compares the time taken by the robot to reach its goal when it is alone to the time it takes when driving in the crowd as: $ T_{r t g}=t_{\text {free }} / t_{c}$, where $t_{\text {free }}$ is the free path completion time, and $t_{c}$ is the robot crowd navigation time. Thus a $T_{rtg}=1$ would be a free path or 100\% efficient navigation.

\textbf{Relative Path length:} The relative path length compares the length of the path taken by the robot to reach its goal with and without a crowd as: $L_{r}=L_{goal} / L_{c}$, where $L_{goal}$ is the shortest path to the goal, and $L_{c}$ is the traveled path with the crowd.

\textbf{Relative Jerk}: The relative jerk evaluates the smoothness of the path taken by the robot to reach its goal ($J_{\text {c }}$) normalized to a reference jerk ($J_{\text {ref }}$) taken during a manual operation of the robot around the crowd as follows:
\begin{IEEEeqnarray}{rCl}
    \begin{aligned}
    & J_{r p}=J_{\text {c }} / J_{\text {ref }}\ , \\
    & J=\left(1 / t_{f}\right) \sum_{t=0}^{t_{f}} \Delta t \sqrt{J_{v}{ }^{2}+J_{\theta}{ }^{2}}\ ,
    \end{aligned}
\label{equ:rel_jerk}
\end{IEEEeqnarray}
where the computed jerk $J$ takes $J_v$ and $J_{\theta}$ from the linear and angular jerk, respectively. $\Delta t$ corresponds to the sample time window over the period $0 \sim t_f$.

\subsubsection{Controller performance}
These tests aimed to estimate the level and type of assistance provided by the reactive controller towards avoiding obstacles. We adapted these metrics from shared control (SC) studies \cite{Carlson2012, Gopinath2016,Erdogan2017}.

\textbf{Contribution}: C is calculated as $C = ||u_r - u_u|| / ||u_u||$, over a finite number of discrete samples N.  Where the $u_r$ is the user command in the robot command space (linear and angular speed), $U_r$ is the output command of the robot. 
Both normalized to the maximum linear and angular velocities. 
In case of full autonomy with a high-level input controller, we evaluated the contribution as $C = ((||u_r ||)/(||u_{u} ||))$ . 
Where $|| x ||$ represents the L2-norm of $x$. 

\textbf{Fluency:} as a measurement of the commands temporal continuity given that the reactive navigation is intervening with the main driving DS desired motion as:
\begin{IEEEeqnarray}{rCl}
    F=\frac{1}{N} \sum_{t=t_{0}}^{t_{N}} 1-\left|u_{h}^{t}-u_{h}^{t-1}\right| ,
    \label{equ:fluency}
\end{IEEEeqnarray}
where $u_h$ is the high-level input from the user. 

\textbf{Agreement}: We defined in terms of the deviation of the direction of the high-level commands from the direction of the final control’s velocity, as follows:
\begin{IEEEeqnarray}{rCl}
    \begin{aligned}
    &\theta(u)=\tan ^{-1}\left(\frac{v}{w}\right), \\
    &a_{i}=1-\left|\theta\left(z_{u}^{i}\right) \ominus \theta\left(u_{S C}^{i}\right)\right| / \pi, \\
    &\text {agreement}=\sum_{i=0}^{N} a_{i} \cdot \Delta t_{i} / \sum_{i=0}^{N} \Delta t_{i},
    \end{aligned}
\label{equ:agreement}
\end{IEEEeqnarray}
where $v$ and $w$ are the linear and angular velocities $u=[v, w]$, $a_i$ is the normalised agreement at time step $t_i$ and $u_{SC}^i$ is the final output of the robot. 
$N$ is the number of samples available in which data from the high-level command input $z_u^i$  coincide in time with $u_{SC}^i$ , and $\Delta t_i$ is the duration of the high-level input command $z_u^i$.

\subsubsection{Pedestrian interaction metrics:}
\textbf{Crowd density:} We report the crowd density in terms of mean, standard deviation and max density. However, the limited field of view of the robot as it is embedded in the crowd hinders overall measurements. Therefore, we reported on the density within 2.5, 5 and 10m around the robot (as shown in Fig. \ref{fig:density}), which explains better the scenario.

\textbf{Minimal distance:} the minimal distance to the robot held by any pedestrian.

\textbf{Virtual Collisions}: number of violations to the virtual boundary set to the robot controller, as shown in Fig. \ref{fig:robot_setup}.

\textbf{Collisions}: number of collisions in a scenario reported by the experimenters and post-analysis of the recordings.

\subsection{Experimental Setup}

\begin{figure}[!t]
    \centering
    \includegraphics[width=8.0cm]{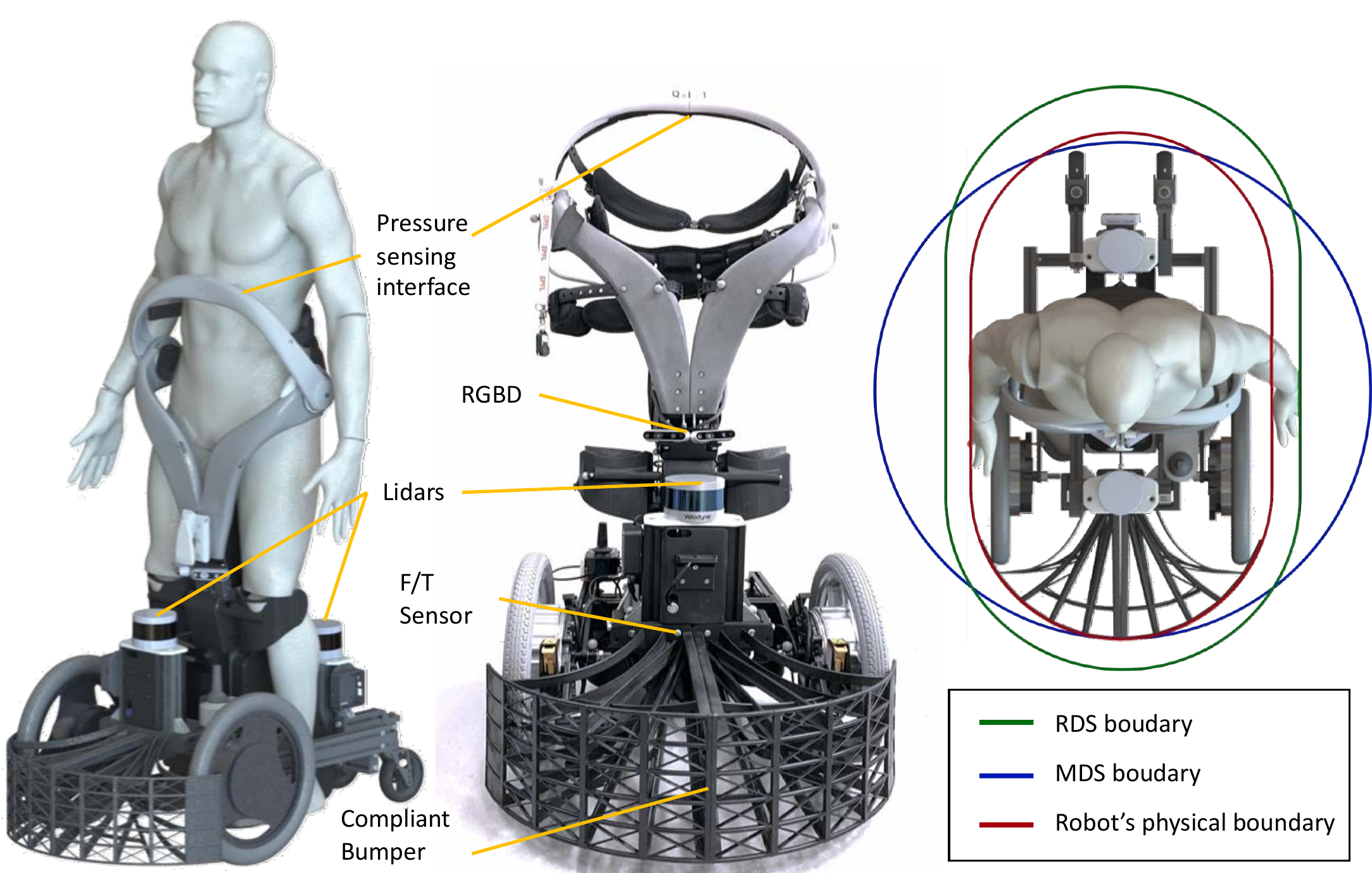}
    \caption{Robot sensing implementation on Qolo \cite{Paez_tmech_2022}: obstacle perception with 2 LIDARs (Velodyne VLP-16), RGBD  sensors (Intel Realsense D435), force sensing with one Rokubi 2.0 (Botasys). With the perception and control implemented on two embedded computers Upboard Squared, and one Nvidia Jetson Xavier AGX. On the right, the boundary representations used around the actual robot
    \label{fig:robot_setup}}
\end{figure}
In this work, we present an evaluation of natural crowd navigation through reactive control. We have used the robot Qolo \cite{Paez_tmech_2022} - a person carrier mobile robot for lower-limb impaired users - instrumented as shown in Fig. \ref{fig:robot_setup}.

We present a comparative evaluation of two modes of obstacle avoidance (MDS and RDS) driven by a linear DS towards a set attractor (or goal some distance from the starting point). And compare it to a baseline of shared-control where a user guides the high-level interaction with the crowd (effectively, deciding the direction of motion).
Such evaluation allows observing the compatibility of the methods with the crowd, the high-level commands, and the effectiveness of pure reactive navigation.
Two navigation scenarios were developed for different crowd densities:
\subsubsection{Scenario 1} 
In this scenario, we performed crowd navigation on a mixed-traffic street  (Rue de Saint-Laurent, Lausanne). The path was chosen at an intersection of 6 streets (as shown in Fig. \ref{fig:exp_setup}, top).
We encountered flows from low to mid-density with mixed crowd types: 2D flows and sparse crowds with static pedestrians.
The sparse crowds' mean density was between $0.05\ ppsm$ to $0.15\ ppsm$ (40 records 50 m round trips, with 33 successful trajectories) and a max crowd density of $0.7\ ppsm$.
The experiment was repeated over multiple dates during a farmer's market to elicit similar crowds.

\subsubsection{Scenario 2} 
\begin{figure}[!t]
    \centering
    \includegraphics[width=8.0cm]{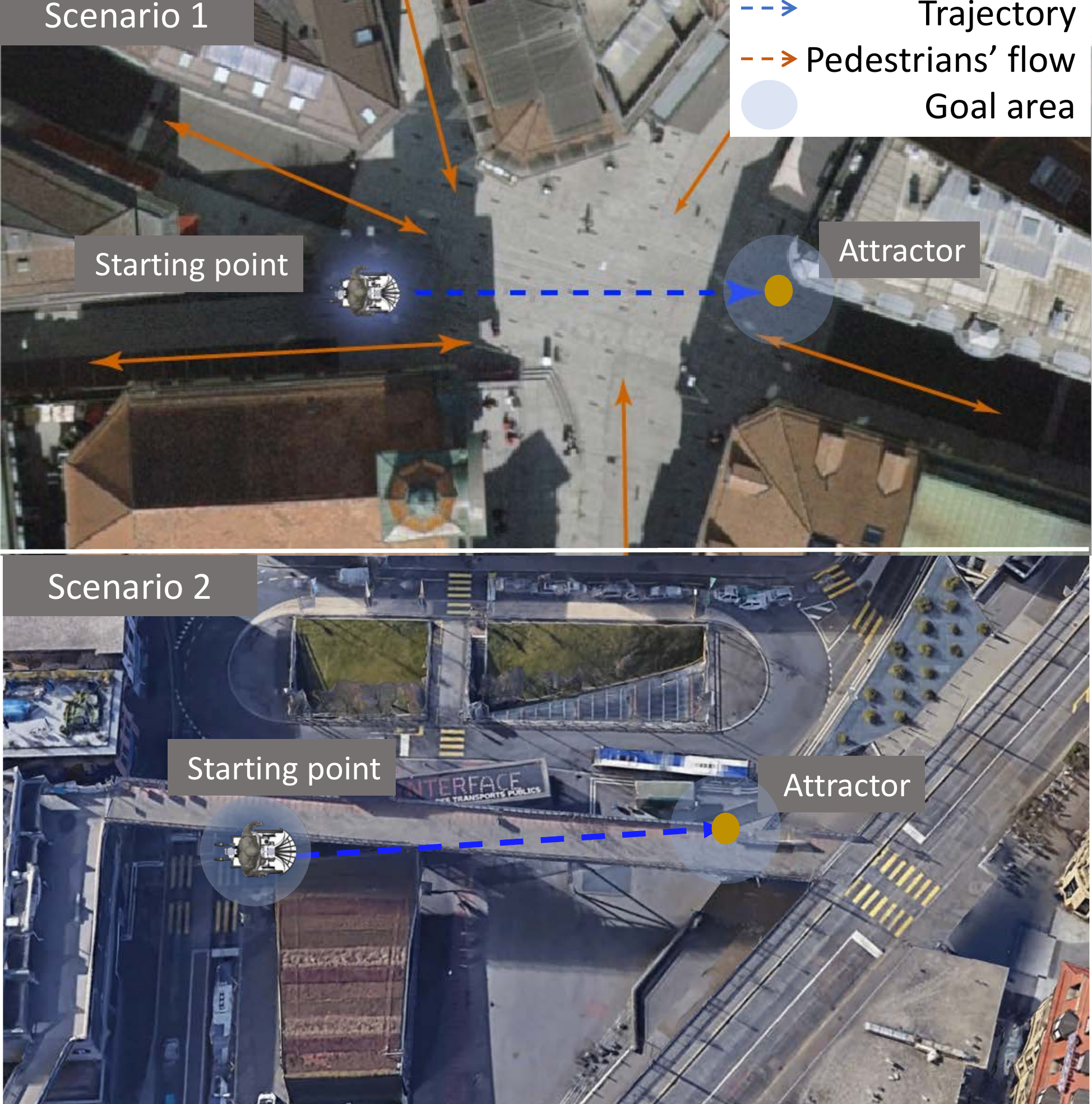}
    \caption{Experimental setup on two mixed traffic streets with a fixed starting point and goal (attractor). Scenario 1, constitutes a mixed influx of 6 streets during a farmer's market with up to $0.7\ ppsm$. 
    Scenario 2, is a closed pedestrian street with crowd crowds up to $1\ ppsm$, and the corresponding experiment was conducted during the Christmas market.
    \label{fig:exp_setup}}
\end{figure}

In this scenario, we performed 1D flow navigation during the Christmas market (Place de l'Europe, Lausanne), as depicted in Fig. \ref{fig:exp_setup}, bottom. 
Here, we found denser mixed crowds formed by people lining up and flows of pedestrians. 
In this case, we followed the same protocol of traversing a set path whose starting and ending goals were 30 m apart.

All data were recorded on ROS standard data type (rosbag) with the following information:
\begin{itemize}
    \item Pedestrian motion information in the form of a set of two sets of 3D point clouds around the robot, including all surrounding people and obstacles in a range of up to 50 m.
    \item Pedestrian's motion data from a forward-looking RGBD camera, with people, labelled and blurred.
    \item Output from 3 people detection layers and 1 integrated people tracker.
    \item Force/Torque information gathered by the contact sensors at the robot's bumper. 
    \item Recordings of the navigation interface input given by the user/driver of the robot.
    \item Motion data was gathered from the robot inertia sensors and odometry sensors.
    \item Video recording of the scene from the robot's perspective without personal identification recording.
    \item *Video recording of the participant-robot driving in a scenario, with no personal identification.
\end{itemize}
All data captured have been released as an open-access dataset \cite{Paez_CrowdDataset_2021}, with rosbag files and post-processing data for each of the following scenarios.

\section{Results} \label{sec:results}

\begin{figure*}[!t]
     \centering
         \subfigure[Snapshot of the experiment where a user onboard the robot Qolo acts as safety operator while the robot drives in full autonomous mode to navigate 20 m ahead in the mixed crowd flow with parts of the trajectory in low-density $<0.2 ppsm$ ($t=2, 12, 38$), and mid-density $<0.7 ppsm$ ($t=24$). ]{\includegraphics[width=17.0cm]{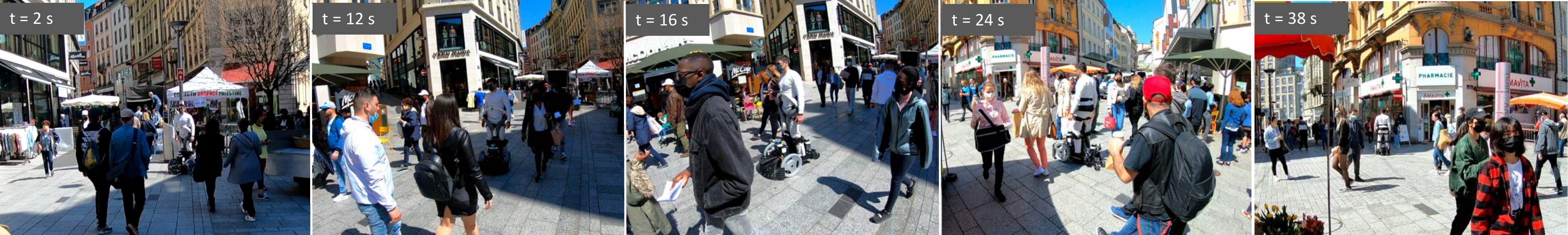}%
    		\label{fig:timelapse}}
		\hfil 
		\subfigure[Robot trajectory over one round trip to the goal location (further point to the right). 
		Detected pedestrians are over-imposed with labelled numbers and corresponding colours at t=5, t=75, and t=118.6.
		]{\includegraphics[height=4.8cm]{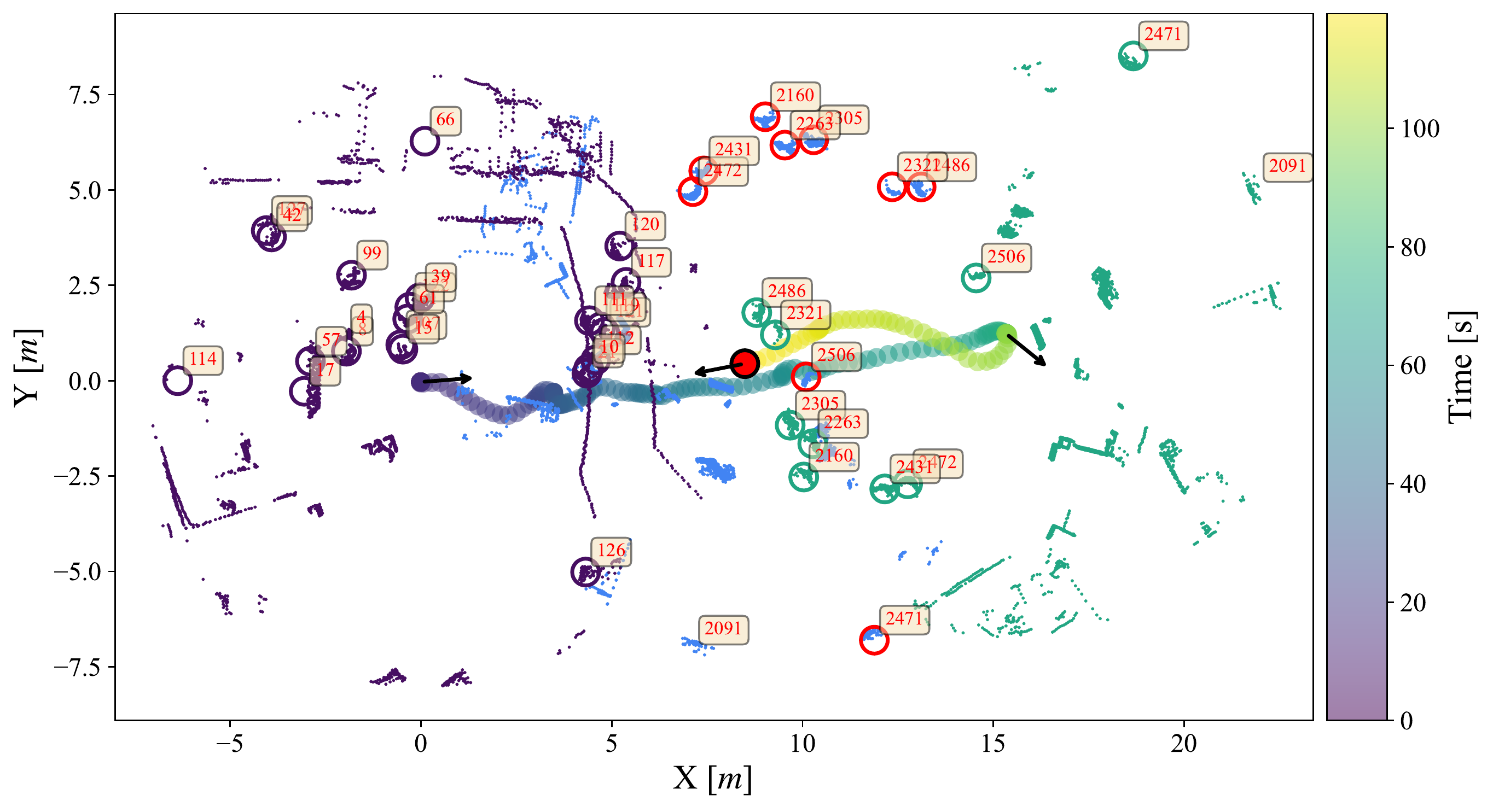}%
		\label{fig:traject}}
		\hfil 
    	\subfigure[Crowd density estimated from online sensing of people detection \cite{Jia2020} at 2.5 m, 5.0 m, and 10 m around the robot. The time labels denote the start and end of the first trajectory to reach the goal. The subsequent time is the return to the starting point. ]{\includegraphics[height=4.8cm]
    	{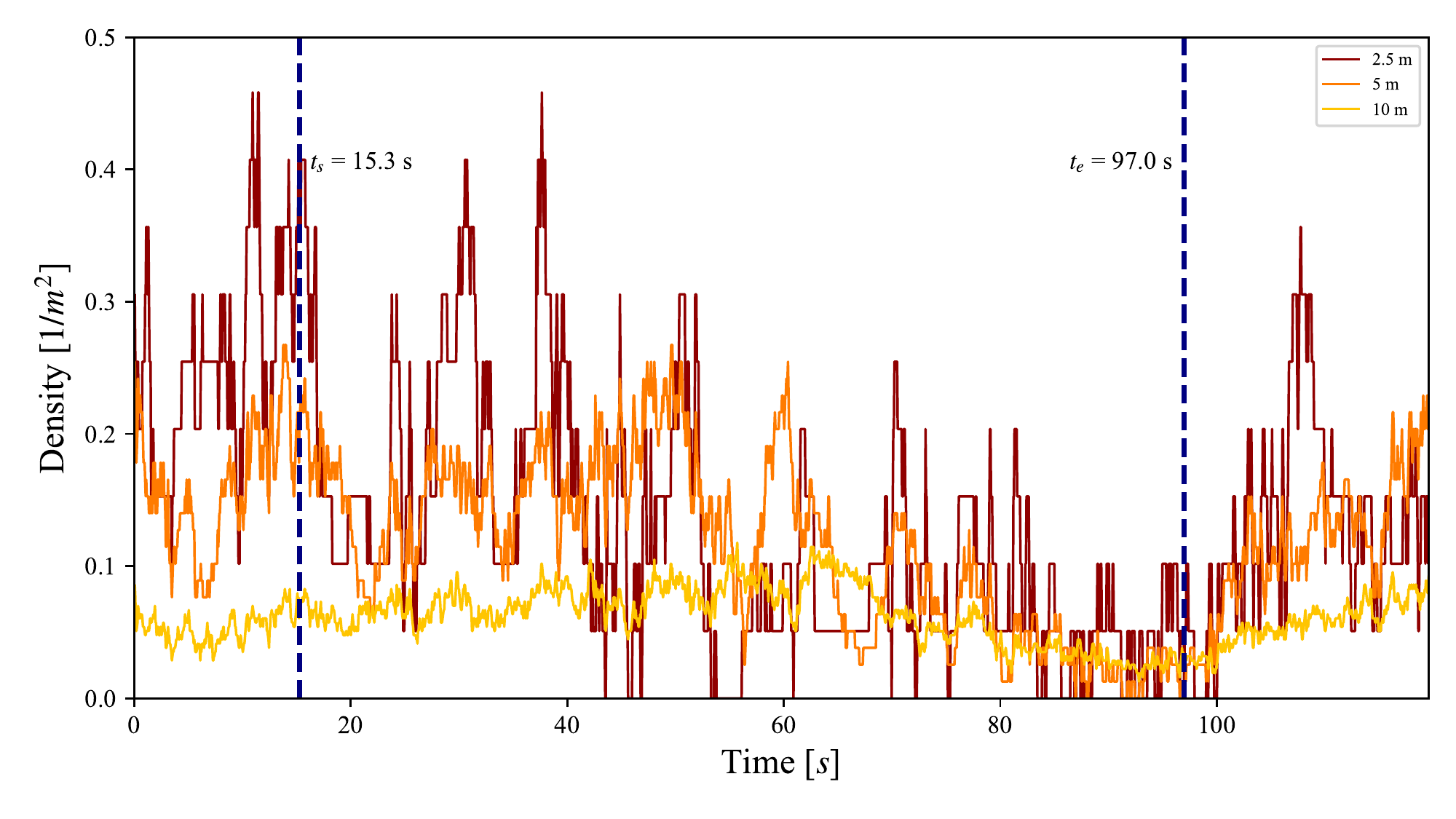}
		\label{fig:density}}
		\hfil
		\caption{Experimental setup example of one of the 110 trials of point-to-point navigation running around crowds in the city of Lausanne (Switzerland)). }\label{fig:exp_sample}
\end{figure*}

Scenario 1 was successfully recorded 95 times for about 5.0 km in sets of 20 m round trips (with 3 interrupted trials). A sample data is given in Fig.\ref{fig:exp_sample} around low and mid crowd densities.
Scenario 2 was successfully recorded 15 times with the highest densities close to $1\ ppsm$ from on-board measurements.

\subsection{Controller Response in Crowds}
The overall crowd density (as shown in Fig. \ref{fig:density}) measured from the onboard sensing fluctuates around $0.12\ ppsm$, with peaks of up to $0.6\ ppsm$. Although it is worth noting that the current measurements are limited to the proximity sensors, therefore, it could be possible that higher densities were not visible from the robot's viewpoint.
Out of successfully recorded 16 MDS trials, 28 RDS trials, and 48 SC trials. 
We qualified all data of successfully reaching the goal with a margin of 3 m around the goal (as depicted in Fig. \ref{fig:exp_setup}). Moreover, we took compatible trials with no significant difference in crowd density variance among tests (mean and max).
Resulting on a set of 16 MDS trials, 20 RDS trials, and 16 SC trials for comparison. 

\begin{figure}[!t]
     \centering
         \subfigure[Controller contribution. ]{\includegraphics[width=4.2cm]{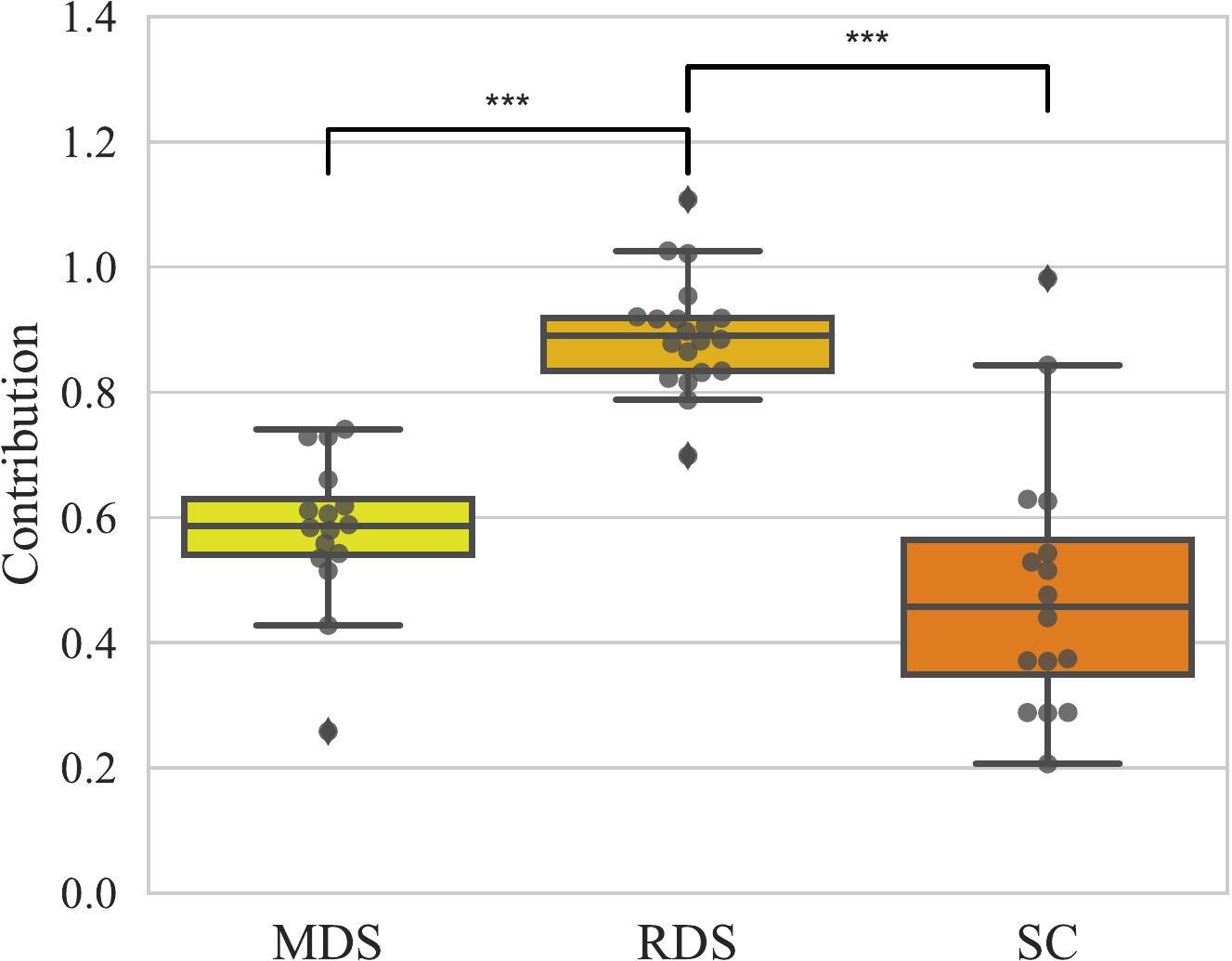}%
    	\label{fig:cntr_contribution}}
		\hfil 
		\subfigure[Controller agreement. ]{\includegraphics[width=4.2cm]{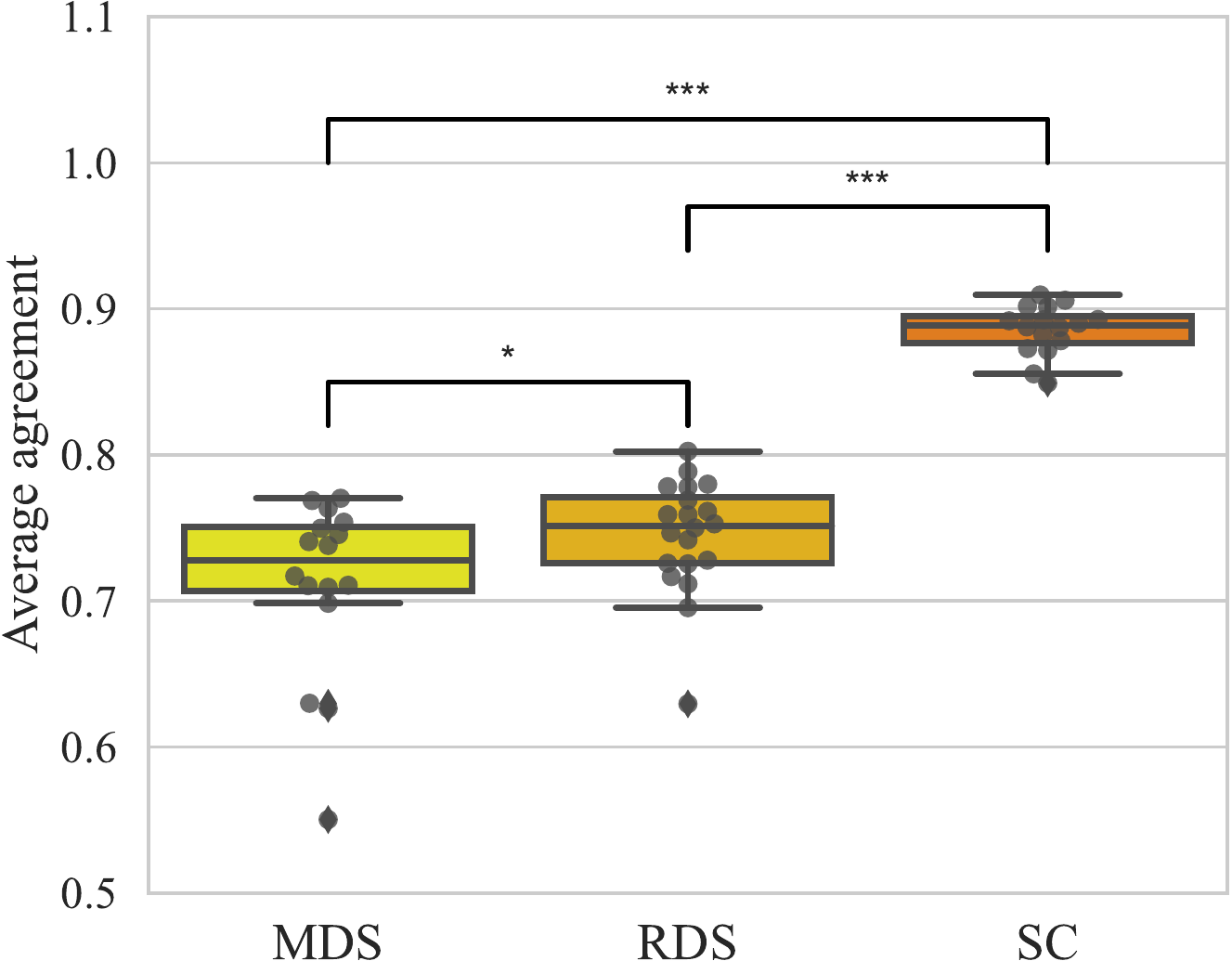}%
    		\label{fig:cntr_agree}}
		\hfil
        \subfigure[Controller fluency. ]{\includegraphics[width=4.2cm]{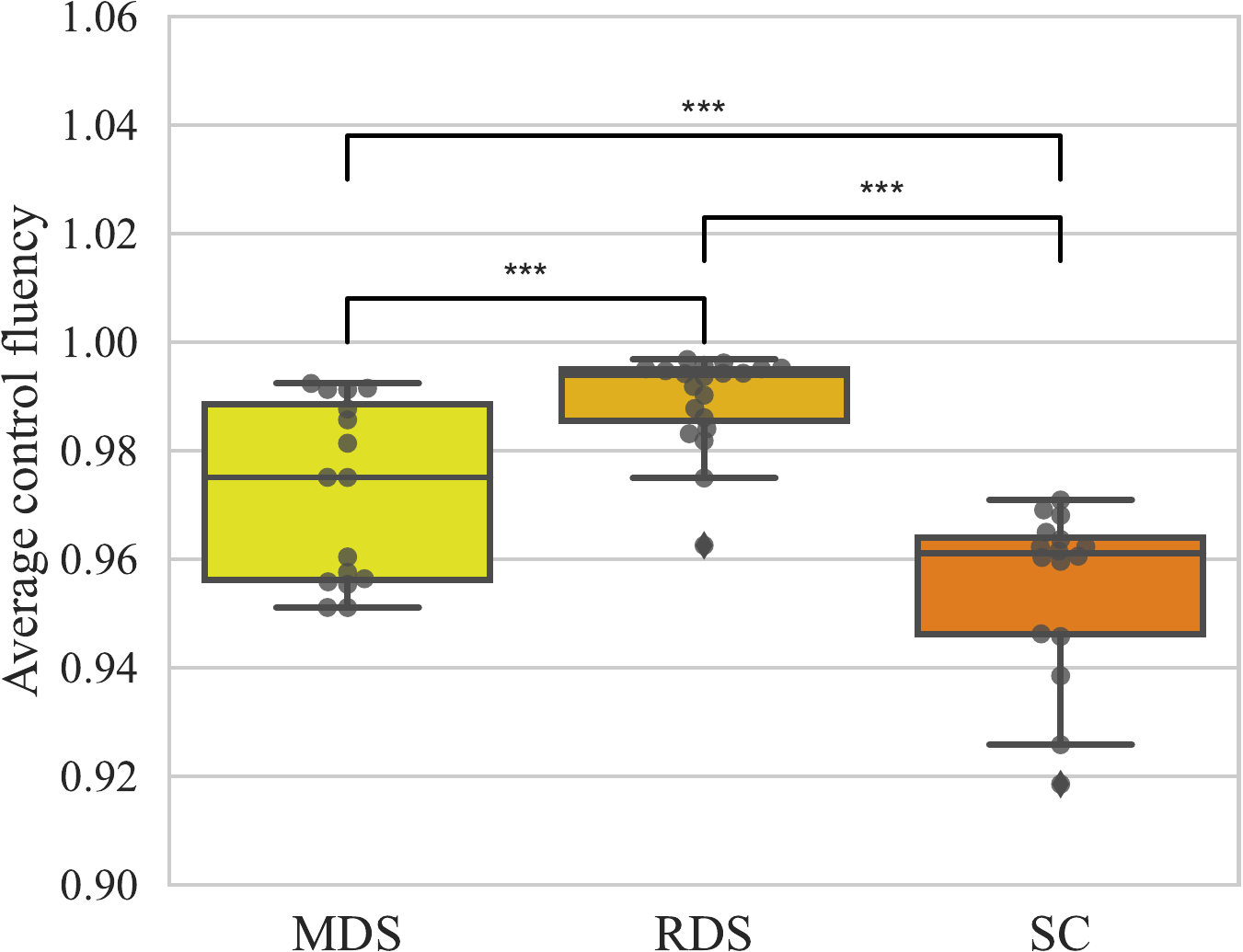}%
    		\label{fig:cntr_fluency}}
		\hfil
 		\subfigure[Relative Jerk. ]{\includegraphics[width=4.2cm]{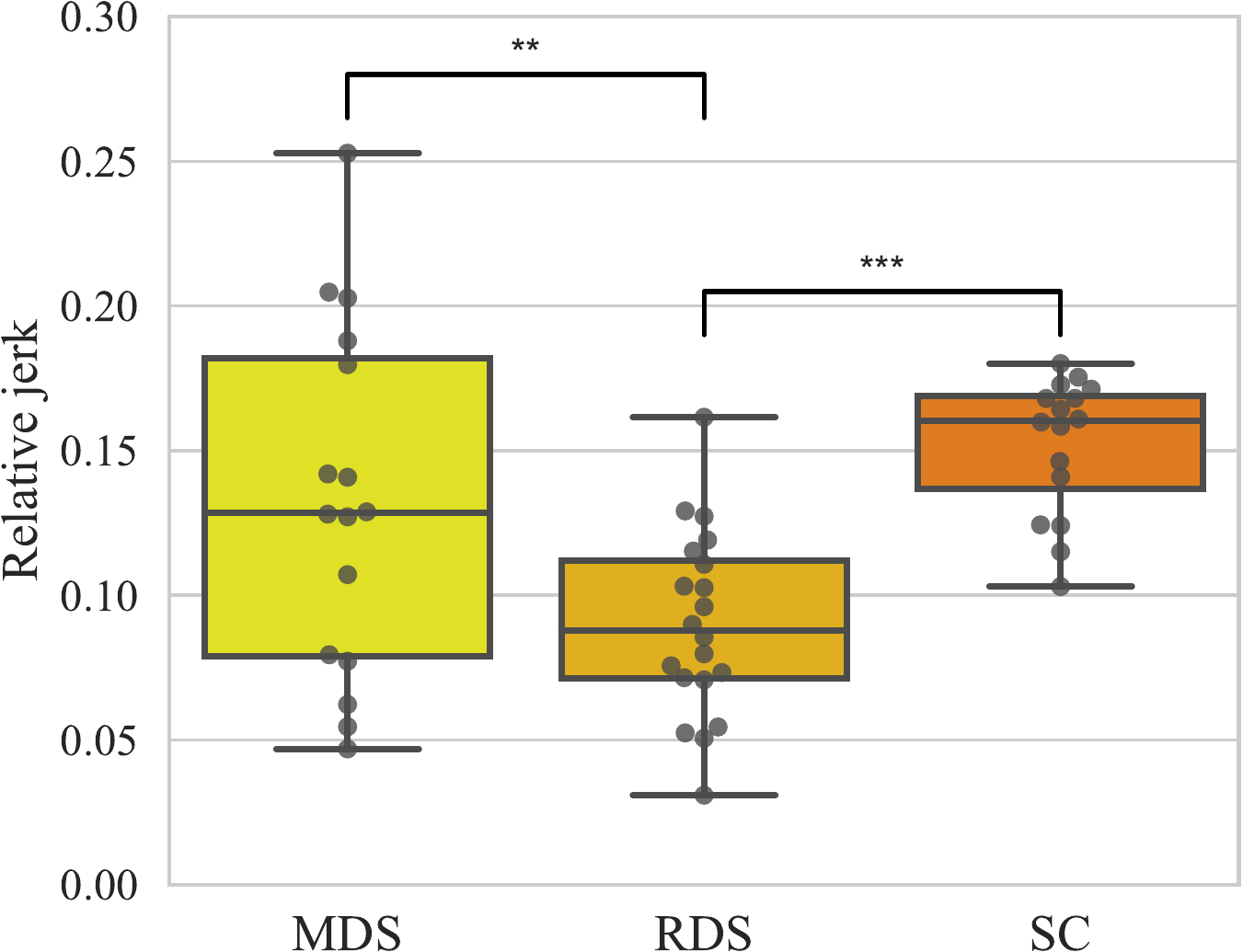}%
    		\label{fig:cntr_jerk}}
		\hfil
		\caption{Controllers evaluation among the task on scenario 1 on crowds between low-mid densities (up to 0.6 ppsm).  All tests were compared by a one-way ANOVA at the levels $* \rightarrow p<0.1$, $** \rightarrow p<0.05$, and $*** \rightarrow p<0.01$. }\label{fig:res_controller}
\end{figure}

In the interaction with the crowd, the minimal distance to pedestrians (Table \ref{tab:control_statistics}) during the trials showed to be lower in the case of the RDS controller, compared with MDS ($p < 0.05$, $F = 3.9$), which is expected for the RDS controller having a more tight definition of the robot's shape. 
Nonetheless, in shared control where the obstacle avoidance is provided by RDS, there was a significantly higher minimal distance ($p < 0.05$, $F=4.4$) to surrounding pedestrians. This suggests that the high-level input from a user would prefer a higher proxemic distance to the pedestrians than the reactive controllers provided ($\sim 0.5 m$).

The path efficiency results show a relative time to the goal similar among controllers, with relative times from 30\% up to 60\% the efficiency of a free path. No significant difference was observed among the methods. This suggests that the reactive navigation for obstacle avoidance does not hinder performance when compared with SC where the user can take lead over the direction of motion.
The relative path length was similar among all recordings of MDS ($21.5 \pm 5.9 m$), RDS ($20.2 \pm 6.2 $), and SC ($27.1 \pm 12.8$ ), showing no significant difference.

The relative jerk (Fig. \ref{fig:cntr_jerk}) showed significant difference between MDS and RDS controllers ($p <0.1, F= 7.3$), with RDS being 5\% lower. RDS and SC mode ($p <0.01, F= 41.8$), with RDS being 6\% lower. 
On the contrary, comparing SC and MDS did not show a significant difference.

In turn, the fluency of the controllers (Fig. \ref{fig:cntr_fluency}) was significantly different among all tested controllers ($p <0.01$, $F = 28.4$), however, only in SC tests was observed a noticeable difference with a 2\% drop in the fluency of the user-robot commands. Overall, this high fluency demonstrates the compatibility of the reactive controllers with the high-level planner.

Analyzing the controller contribution and agreement we observed that RDS control guides most of the linear velocity components with an average 87\% contribution, whereas the MDS controller provided 60\% of contribution (significantly different at the level $p <0.01$, $F = 80.6$) and agreement on the 70\%. 
The difference in the approach to avoiding obstacles is clear, with RDS controlling the magnitude of the velocity, whereas MDS executes higher control over the angular component.

Compared with SC, the results (see, Fig. \ref{fig:res_controller}) show that the obstacle avoidance contributed to 50\%, with agreement over 85\%, which means that the user was in control of the motion direction 50\% of the time, and the obstacle avoidance assisted mostly in the velocity magnitude over the tests. 

\begin{table}[!t]
    \centering
    \scriptsize 
    \footnotesize
    \caption{Controller comparison in mid-density crowds} \label{tab:control_statistics}
    \begin{tabular}{llll}
    \toprule
    \multicolumn{1}{l}{\multirow{2}[4]{*}{\textbf{Metrics}}} & \multicolumn{3}{c}{\textbf{Controller}} \\
    \cmidrule{2-4}          & \multicolumn{1}{l}{\textbf{MDS}} & \multicolumn{1}{l}{\textbf{RDS}} & \multicolumn{1}{l}{\textbf{Shared control}} \\
    \midrule
    Avg. crowd density &  $0.12  \pm 0.03$ &  $0.13  \pm 0.03$ &  $0.12  \pm 0.03$ \\
    Max crowd density &  $0.45  \pm 0.08$ &  $0.47  \pm 0.12$ &  $0.51  \pm 0.14$ \\
    Min distance &  $1.19  \pm 0.16$ &  $1.08  \pm 0.18$ &  $1.20  \pm 0.16$ \\
    Time to goal    &  $0.28  \pm 0.09$ &  $0.32  \pm 0.10$ &  $0.29  \pm 0.07$ \\
    Path length &  $1.41  \pm 0.21$ &  $1.34  \pm 0.20$ &  $1.52  \pm 0.52$ \\
    Jerk             &  $0.13  \pm 0.06$ &  $0.09  \pm 0.03$ &  $0.15  \pm 0.02$ \\
    Contribution         &  $0.58 \pm 0.12$ &  $0.89 \pm 0.09$ &  $0.49 \pm 0.21$ \\
    Avg. fluency          &  $0.97 \pm 0.02$ &  $0.99 \pm 0.01$ &  $0.95 \pm 0.02$ \\
    Avg. agreement        &  $0.71 \pm 0.06$ &  $0.74 \pm 0.04$ &  $0.89 \pm 0.02$ \\
    Virtual collision    &  $3.50 \pm 2.71$ &  $7.05 \pm 7.92$ &  $4.25 \pm 3.11$ \\
    Actual collision    &  2/16 &  2/20 &  3/16 \\
    \bottomrule
    \end{tabular}
\end{table}

\subsubsection{Collisions:}
As detailed in table \ref{tab:control_statistics}, we observed, across all tests, seven collisions leading to contact with pedestrians. 
All of which occurred at the frontal bumper of the robot.
In 5 of the collisions, either the operator on-board the robot or the external supervisor of the tests pressed the emergency stop, therefore no post-collision control was recorded on those occasions.
In the remaining 2 contacts (contacts with the RDS model) the contact was with a pedestrian's shopping trolley, and the contact was below impact threshold $F_c=45N$, so that, there was not any perceived reaction from the compliance controller, nor was collision perceived by the pedestrian.

We documented each collision case and found the following possible causes for the controller behaviour.
First, limited acceleration on the robot that could not match the pedestrian's acceleration. Probable cause of one collision with RDS test, and one with SC.
Second, non-holonomic constraints in dense crowds limit reactivity, a probable cause for one collision with MDS.
Third, pedestrians being very close to a surrounding obstacle generated false negatives in pedestrian detection. Probable cause for two collisions with SC and MDS tests.

Virtual collisions, defined as entering the virtual space of the robot controller (as depicted in Fig. \ref{fig:robot_setup}). We found a slightly lower number of cases on the MDS controller compared with RDS ($p < 0.1$, $F =  2.9$), with no significant difference to SC. This result is expected given the larger size of the robot representation used in MDS, which forces the robot to be further away from pedestrians.

\subsection{Crowd Density Influence on Controller Response}

\begin{figure}[!t]
     \centering
		\subfigure[Max crowd density (2.5m). ]{\includegraphics[width=4.2cm]{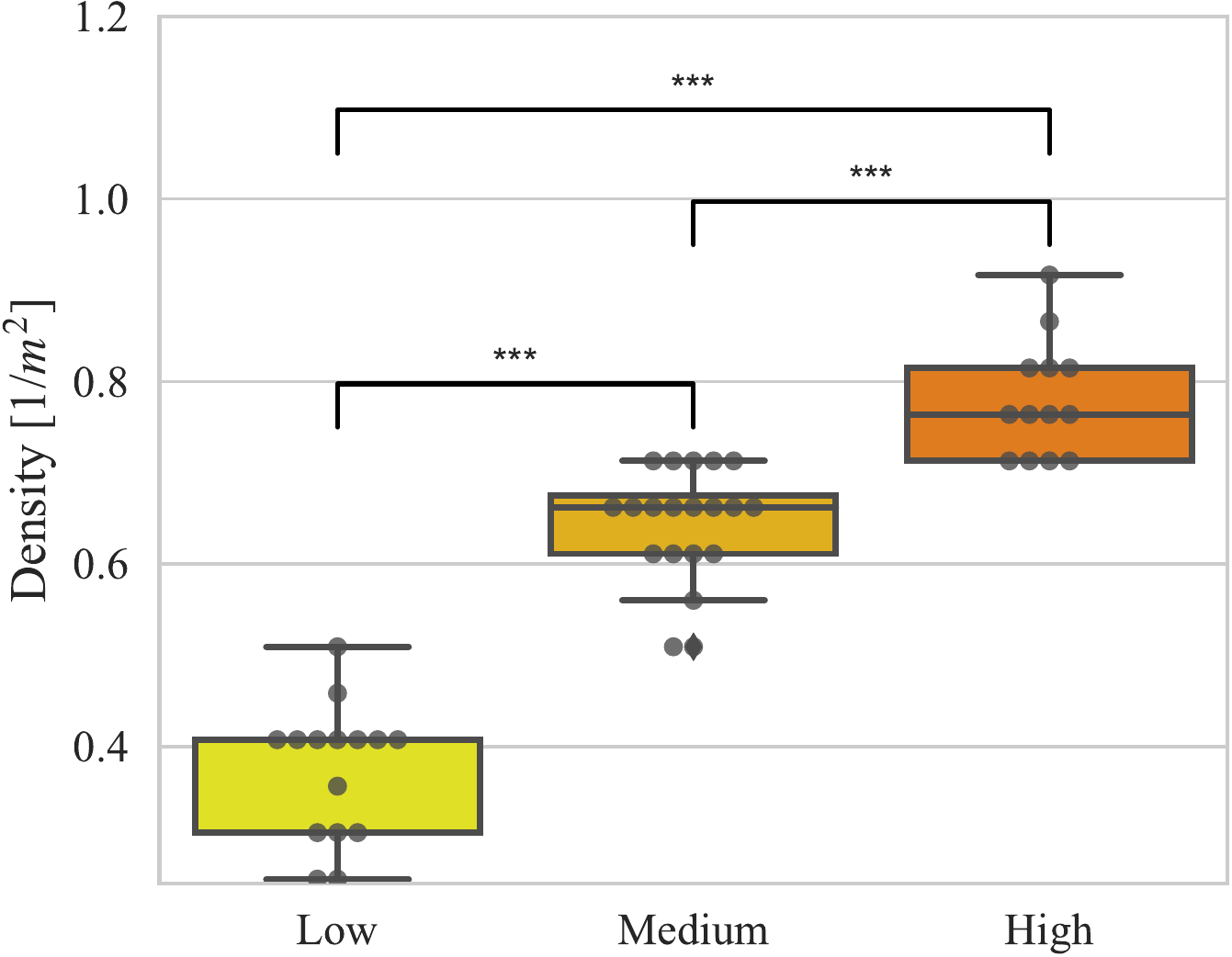}%
    		\label{fig:den_max_den}}
		\hfil
		\subfigure[Virtual collisions (entering the robot's safe space). ]{\includegraphics[width=4.2cm]{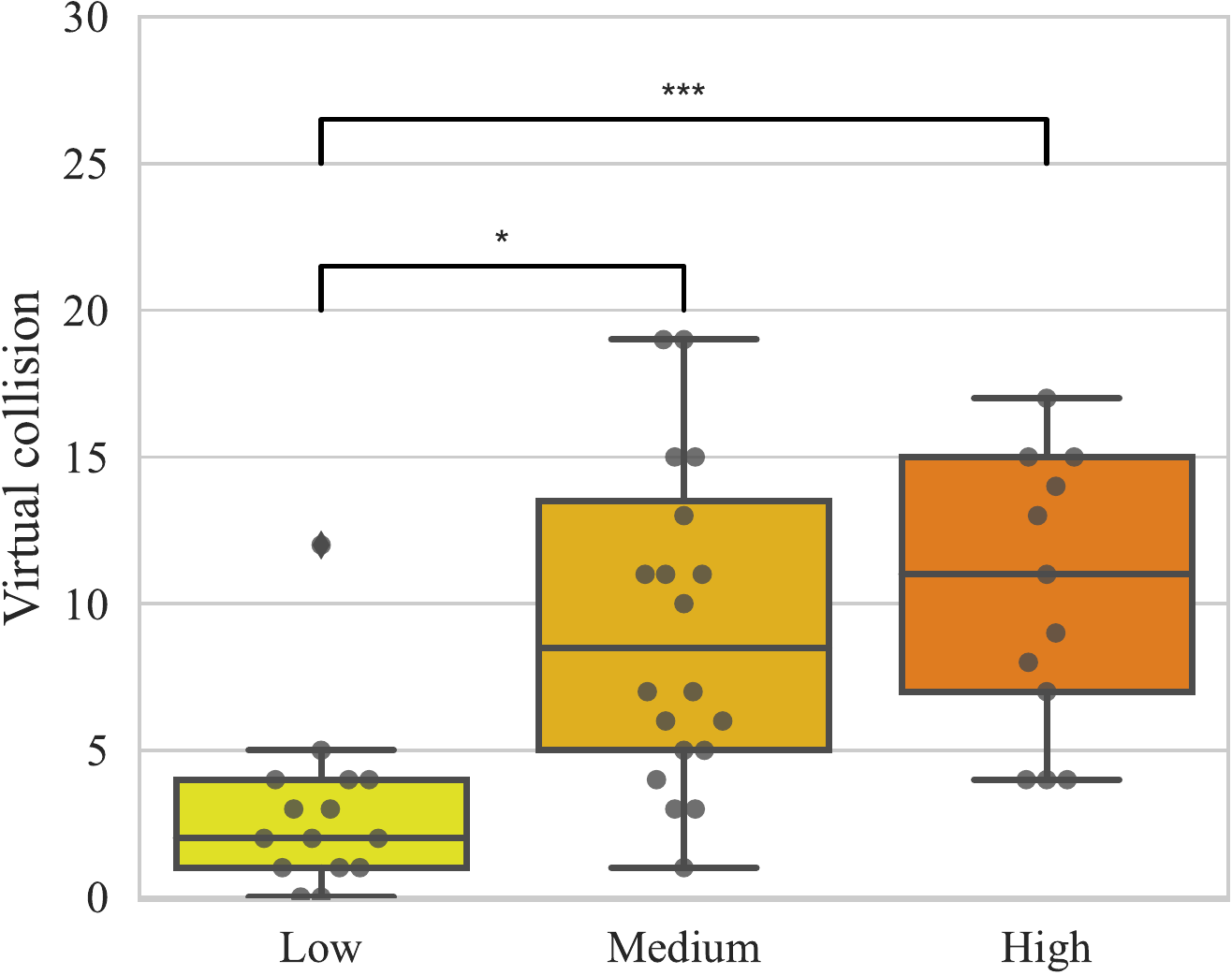}%
    		\label{fig:den_collision}}
		\hfil
		\subfigure[Minimal distance to pedestrians. ]{\includegraphics[width=4.2cm]{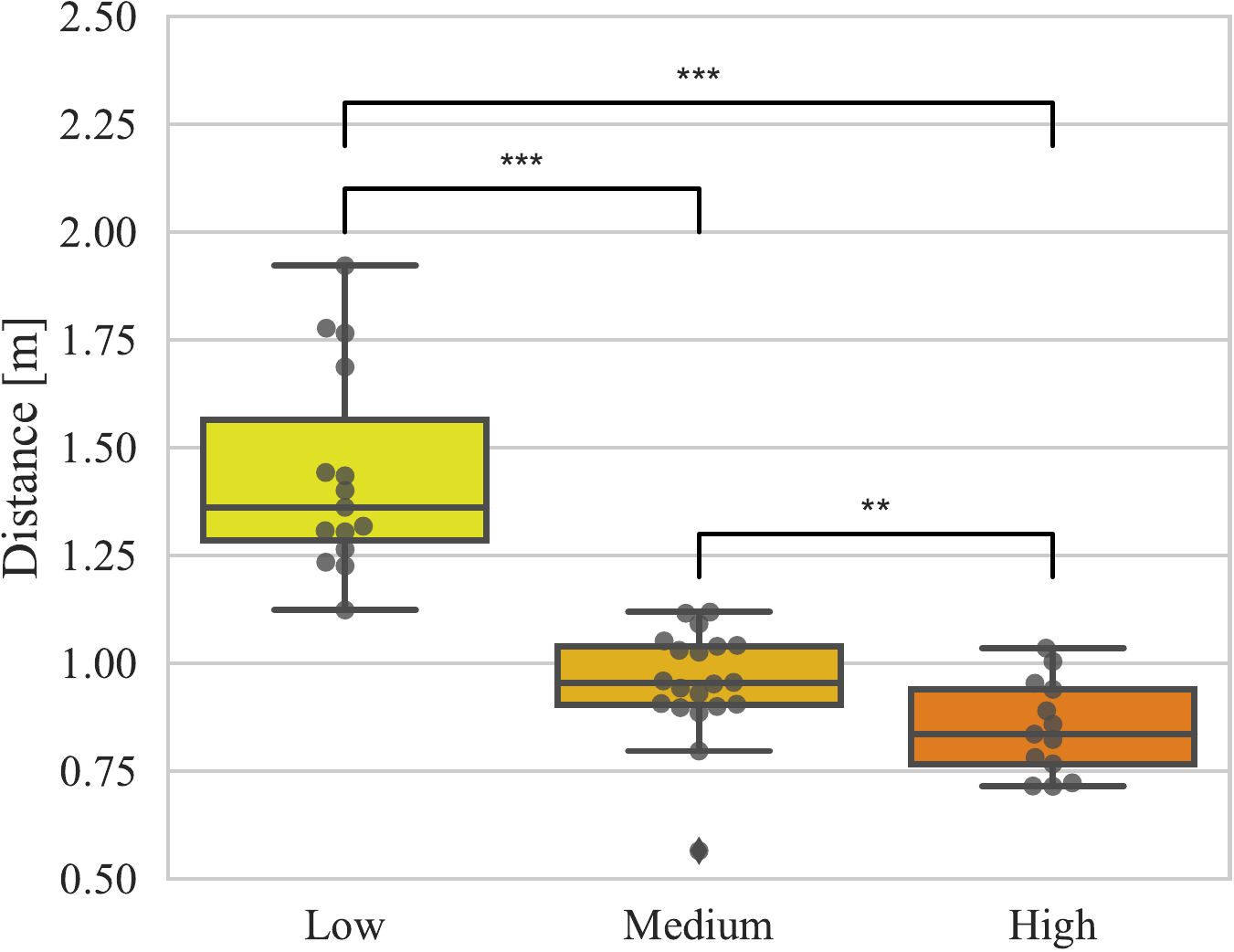}%
    		\label{fig:den_distance}}
		\hfil
		\subfigure[Relative time to goal. ]{\includegraphics[width=4.2cm]{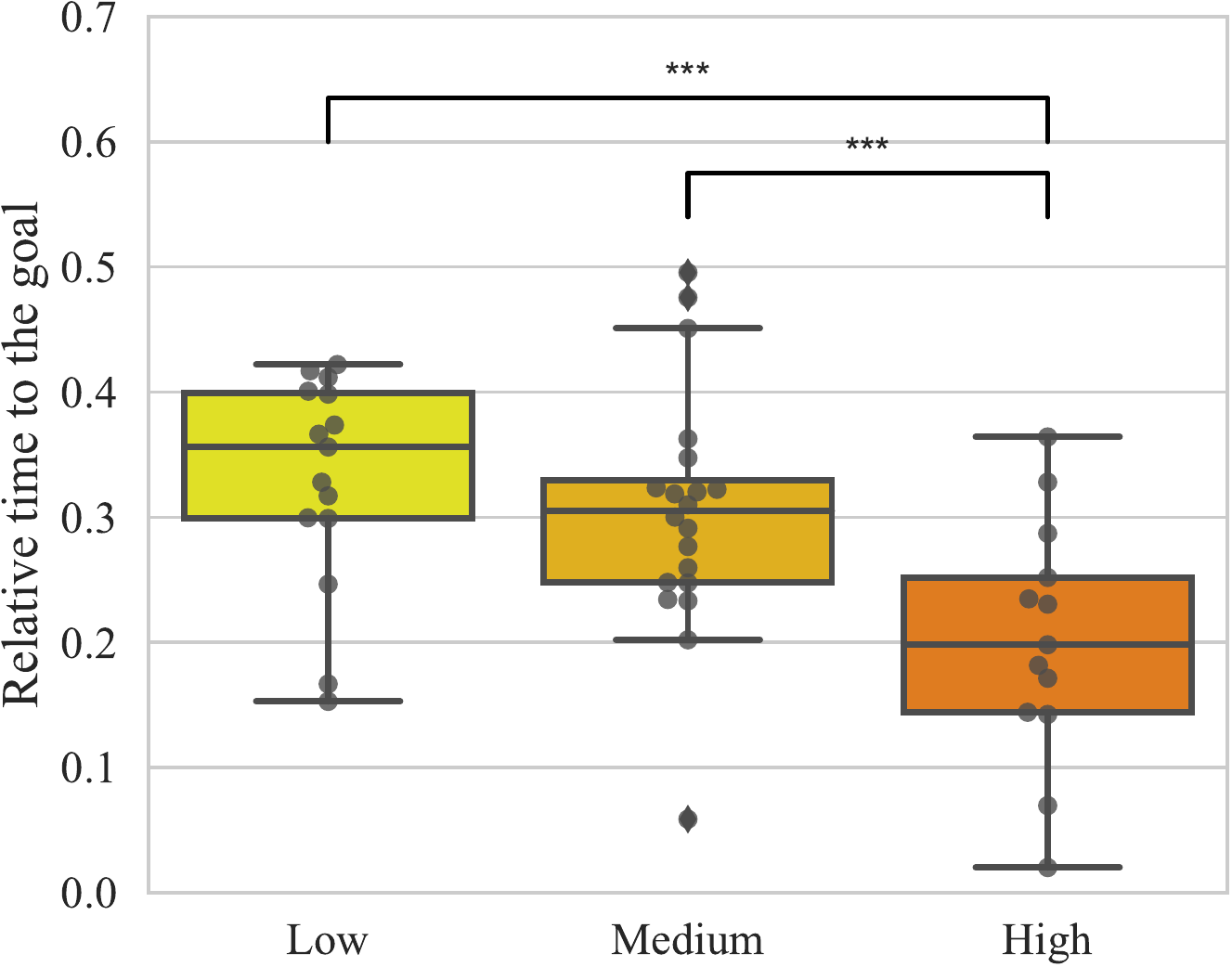}%
    		\label{fig:den_path}}
		\hfil
		\subfigure[Controller contribution. ]{\includegraphics[width=4.2cm]{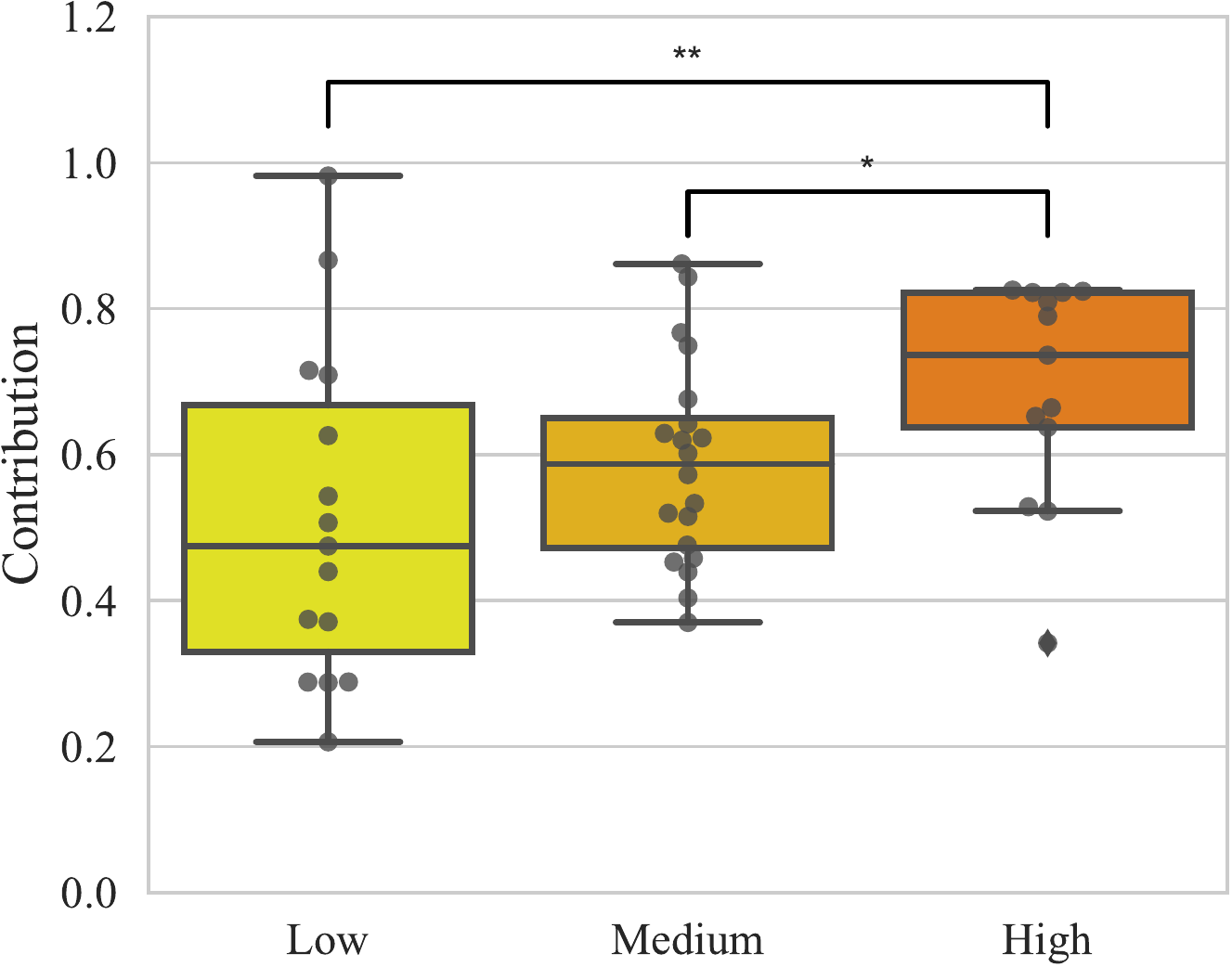}%
    		\label{fig:den_contribution}}
		\hfil
		\subfigure[Controller agreement. ]{\includegraphics[width=4.2cm]{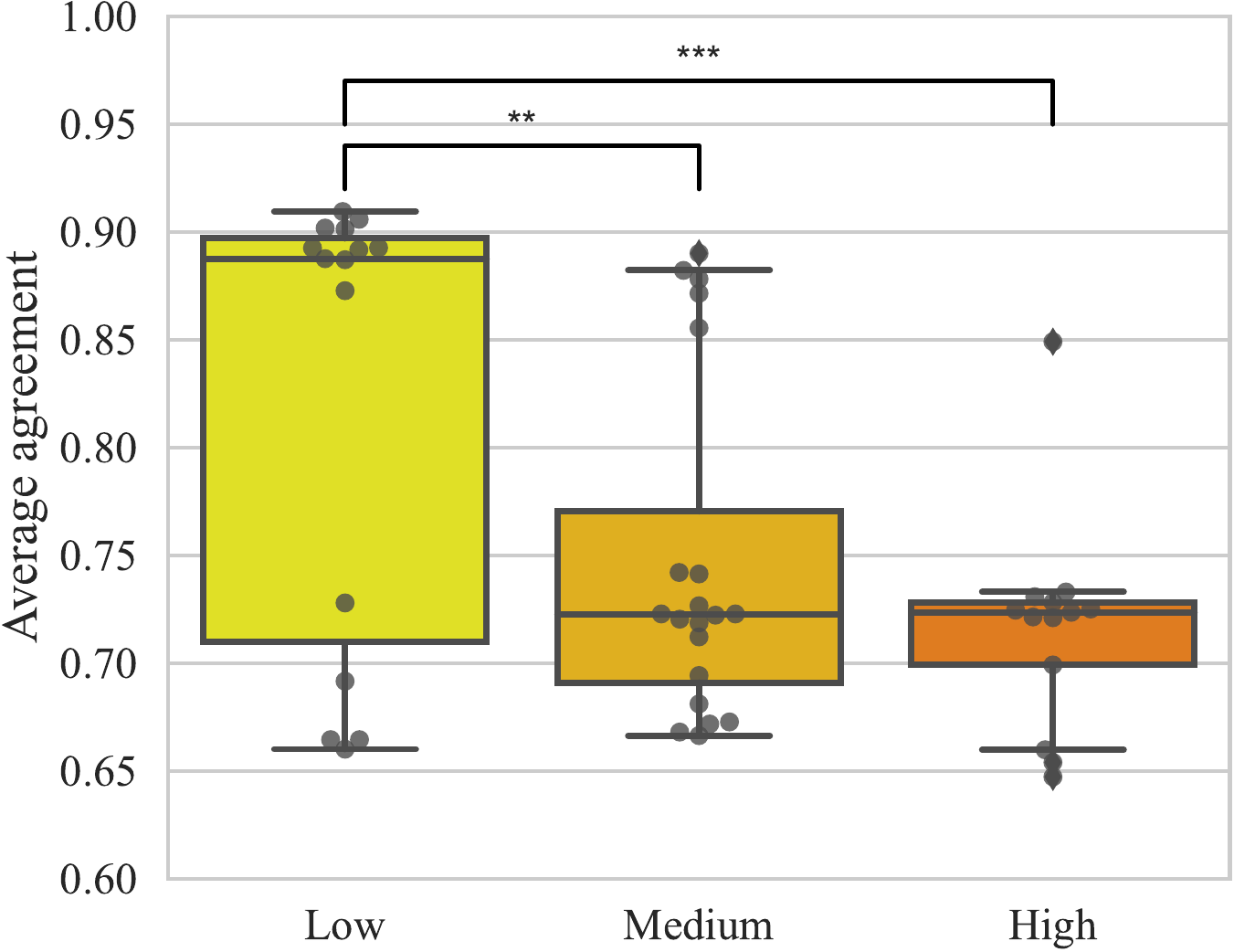}%
    		\label{fig:den_agreement}}
		\hfil
		\caption{Navigation evaluation among the task on scenarios 1 and 2 comparing reactive navigation performance by crowd density. The cluster size was 15, 20, and 13 for low, medium and high density, respectively. All tests were compared by a one-way ANOVA at the levels $* \rightarrow p<0.1$, $** \rightarrow p<0.05$, and $*** \rightarrow p<0.01$. }\label{fig:res_density}
\end{figure}

We took 48 samples from scenarios 1 and 2 of successful crowd navigation within the set points using the shared control navigation.
Subsequently, we cluster the data in 3 groups of low, medium and high densities considering average, variance and max densities in $2.5 m$ and $5 m$, resulting in the metrics computed in table \ref{tab:density_cluster_statistics}.
The metrics with a significant difference among the density conditions are shown in Fig. \ref{fig:res_density}.

The proximity of pedestrians clearly drops consistently with the crowd density ($p<0.01$, $F=53.3$) as shown in Fig. \ref{fig:den_distance}. Meanwhile, the average number of virtual collisions increases significantly only to medium densities ($p<0.01$, $F=15.3$). We observe, in sparse crowds a mean of 2.9 virtual collisions per test, whereas an average of 12 collisions per test in dense crowds.

Nonetheless, it is worth noting that in high-density crowds no actual collisions or contacts were recorded. Contrary to a medium density where we recorded 3 contacts.

Comparing the path efficiency, we found a significant difference ($p<0.01$, $F=8.5$) in the relative time to goal with navigation on high density dropping $10\%$ of the relative time.
In contrast, relative path length showed no significant difference among the tested crowd environments.  Hence, we observe an expected behaviour compatible with human crowds where the reactive controller operates by adjusting its velocity to the crowd.

Fig. \ref{fig:den_contribution} shows the comparison of the controller contribution for the navigation. We observe no significant difference between low-mid densities, with only high density showing a significantly higher contribution ($p <0.05$, $F=3.9$) with a mean of 69\%. 
In the agreement metric with the user, we observe significantly high values (82\%) in low density ($p<0.05$, $F=5.8$), which drops to 75\% in medium density, and 72\% in high density. 
i.e., sparse low-density crowd navigation required less reactive control assistance whereas in the cases of mid- and high-density required increased reactive control contribution although no decreased agreement was registered. 
The user and the controller likely agree on the direction due to the more restrictive available space in the flow of dense crowds

\begin{table}[!t]
    \centering
    \tiny 
    \footnotesize
    \caption{Crowd Navigation Comparison in different densities}
    \label{tab:density_cluster_statistics}
    \begin{tabular}{llll}
    \toprule
    \multicolumn{1}{l}{\multirow{2}[4]{*}{\textbf{Metrics}}} & \multicolumn{3}{c}{\textbf{Crowd density}} \\
    \cmidrule{2-4}          & \multicolumn{1}{l}{\textbf{Low}} & \multicolumn{1}{l}{\textbf{Medium}} & \multicolumn{1}{l}{\textbf{High}} \\
    \midrule
    Avg.Density (2.5m) &  $0.08 \pm0.02$ &    $0.18 \pm0.03$ &   $0.26 \pm0.04$ \\
    Max.Density (2.5m) &  $0.37 \pm0.07$ &    $0.64 \pm0.06$ &   $0.78 \pm0.06$ \\
    Avg.Density (5m) &  $0.09 \pm0.02$ &    $0.17 \pm0.02$ &   $0.19 \pm0.02$ \\
    Min.Distance &  $1.44 \pm0.24$ &    $0.96 \pm0.13$ &   $0.85 \pm0.11$ \\
    Time to goal    &  $0.33 \pm0.09$ &    $0.30 \pm0.10$ &   $0.20 \pm0.10$ \\
    Path length &   $1.46 \pm0.57$ &    $1.74 \pm1.28$ &   $2.89 \pm4.07$ \\
    Jerk             &  $0.14 \pm0.03$ &    $0.14 \pm0.03$ &   $0.15 \pm0.03$ \\
    Contribution         &  $0.51 \pm0.23$ &    $0.59 \pm0.14$ &   $0.69 \pm0.15$ \\
    Avg. fluency          &  $0.95 \pm0.02$ &    $0.96 \pm0.02$ &   $0.95 \pm0.01$ \\
    Avg. agreement        &  $0.82 \pm0.11$ &    $0.75 \pm0.08$ &   $0.72 \pm0.05$ \\
    Virtual collision    &  $2.93 \pm2.94$ &  $13.15 \pm19.28$ &  $12.08 \pm8.50$ \\
    \bottomrule
    \end{tabular}
\end{table}

\section{Summary and Conclusions} \label{sec:Discussion}

In the current evaluation of a reactive controller for crowd navigation, we have shown its feasible to navigate around natural pedestrian crowds demonstrating the controller’s compatibility with the behaviour of bystander pedestrians. 
During all 110 trials in different crowd types: sparse, flows, and mixed crowds, we did not report any incidents with pedestrians, although, we recorded 7 contacts within safe limits. 
On the contrary, any comments made by bystanders were positive about the perceived utility of assistive navigation technology for mobility-impaired users.

The reactive navigation controllers showed to perform similarly in trajectory efficiency and fluency during the motions. The controllers differ in the way to avoid obstacles, one mainly deviating through angular velocity (MDS) and the other mostly by reducing speed (RDS). It is hence expected that they lead to measurable differences such as faster displacement to the goal for MDS than RDS (confirmed by results), larger jerkiness for MDS than RDS (confirmed by data). One may have expected that MDS leads to more collisions than RDS but this is not confirmed by the results. However, the number of collisions were too few to conclude.

A significant difference found in the reactive controller response was given by the type of crowd, where sparse and mixed traffic crowds in the medium density showed to be more difficult to navigate with a lower agreement with the user at the same level of contribution. In contrast, flow crowds made it easier for the controller to respond to a somewhat uniform response of the surrounding pedestrians, as validated with the equal agreement levels (see, Fig. \ref{fig:res_density}).

The proposed reactive controller focused on a single layer of the robot response for crowd navigation which performs efficiently compared with shared control, with no significant difference in time to goal, minimal distance to pedestrians, motion jerk, and path length.
Further work should focus on situational awareness of the complex mixed crowd environments, as required when interacting in complex environments, e.g., on some occasions, the robot did not interact appropriately with the environment. i.e., getting inside stores, getting in the middle of people lining up, and entering shops in the wrong direction. In comparison, shared control was more natural, and most of the tests ran smoothly without bystanders even noticing the robot.


\section*{Acknowledgement}
The experiments were approved with an ethical protocol by the human research ethical committee of EPFL (Approval No: HREC-032-2019), and with the approval and cooperation of the office for mobility, the police and the office for parks, and the public domain of the city of Lausanne. Approval numbers: 395128 and 416008.
\textit{Disclaimer:} D.P. and K.S. hold the patents of the robot Qolo and shares in the company Qolo Inc.


%

\bibliographystyle{IEEEtran}
\bibliography{IEEEabrv,crowd_robot_navigation.bib}

\end{document}